\documentclass{egpubl}
\usepackage{eg2021}
 
\ConferencePaper        % uncomment for (final) Conference Paper
\usepackage[T1]{fontenc}
\usepackage{dfadobe}  

\usepackage{cite}  % comment out for biblatex with backend=biber 
\BibtexOrBiblatex
\electronicVersion
\PrintedOrElectronic

\ifpdf \usepackage[pdftex]{graphicx} \pdfcompresslevel=9
\else \usepackage[dvips]{graphicx} \fi

\usepackage{egweblnk} 
\usepackage{setspace}
\usepackage{lipsum}
\usepackage{multirow}
\usepackage{xcolor}
\usepackage{amsmath}
\definecolor{FIXMECOLOR}{rgb}{1,0,0}

\definecolor{CMCOLOR}{rgb}{0,0.8,0}

\definecolor{CMTCOLOR}{rgb}{0.6,0,0.6}

\title[Walk2Map: Extracting Floor Plans from Indoor Walk Trajectories]%
      {{\em Walk2Map}: Extracting Floor Plans from Indoor Walk Trajectories\vspace{-1em}}

\author[C. Mura et al.]
{\parbox{\textwidth}{\centering 
        Claudio Mura$^{1,2,3}$ $\quad$
        Renato Pajarola$^{1}$ $\quad$
        Konrad Schindler$^{2}$ $\quad$
        Niloy Mitra$^{3,4}$ $\quad$
        }
        \\
{\parbox{\textwidth}{\centering $^1$Visualization and MultiMedia Lab, University of Zurich, Switzerland \hspace{2em}
         $^2$Photogrammetry and Remote Sensing, ETH Z\"urich, Switzerland \\
         $^3$Smart Geometry Processing Group, University College London \hspace{2em}
         $^4$Adobe Research London
      }\vspace{-2em}
}
}

\begin{document}

\teaser{
  \includegraphics[width=\textwidth]{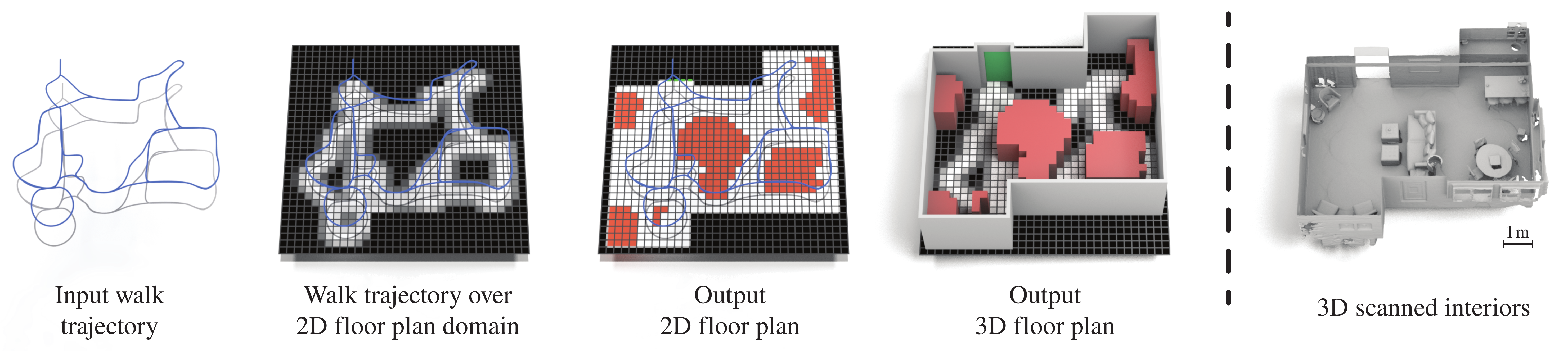}
  \centering
  \caption{We present \emph{Walk2Map}. Starting from \emph{only} a walk trajectory signal in an unknown room, we pose the question: \emph{Can one infer the floor plan in terms of room boundary, door locations, and furniture regions?} The scan on the right shows the actual reference scene. }
  \label{fig:teaser}
}

\maketitle
%-------------------------------------------------------------------------
\begin{abstract}
Recent years have seen a proliferation of new digital products for the efficient management of indoor spaces, with important applications like emergency management, virtual property showcasing and interior design.
While highly innovative and effective, these products rely on accurate 3D models of the environments considered, including information on both architectural and non-permanent elements. These models must be created from measured data such as RGB-D images or 3D point clouds, whose capture and consolidation involves lengthy data workflows. This strongly limits the rate at which 3D models can be produced, preventing the adoption of many digital services for indoor space management.

We provide a radical alternative to such data-intensive procedures by presenting {\em Walk2Map}, a data-driven approach to generate floor plans \textit{only} from trajectories of a person walking inside the rooms. 
Thanks to recent advances in data-driven inertial odometry, such minimalistic input data can be acquired from the IMU readings of consumer-level smartphones, which allows for an effortless and scalable mapping of real-world indoor spaces.
Our work is based on learning the latent relation between an indoor walk trajectory and the information represented in a floor plan: interior space footprint, portals, and furniture.
We distinguish between recovering area-related (interior footprint, furniture) and wall-related (doors) information and use two different neural architectures for the two tasks: an image-based Encoder-Decoder and a Graph Convolutional Network, respectively. We train our networks using scanned 3D indoor models and apply them in a cascaded fashion on an indoor walk trajectory at inference time.

We perform a qualitative and quantitative evaluation using both trajectories simulated from scanned models of interiors and measured, real-world trajectories, and compare against a baseline method for image-to-image translation. The experiments confirm that our technique is viable and allows recovering reliable floor plans from minimal walk trajectory data.
%-------------------------------------------------------------------------




\end{abstract}  

%-------------------------------------------------------------------------

\section{Introduction}
\label{sec:introduction}

The way we organize and interact with indoor environments has a profound influence on many aspects of human activity, from the economic success of large companies to the effectiveness of everyone's daily routine. 
For this reason, significant efforts have been devoted to exploit the latest digital technologies for the optimization of real-world workflows focused on building interiors, such as virtual property showcasing, interior design, indoor space optimization and emergency management. 
A number of innovative services and online platforms have been built around these use cases, including Matterport~\cite{Matterport} and Google Indoor Maps~\cite{GoogleIndoorMaps}, which promise to revolutionize the way we manage and experience indoor spaces.
Such products rely on the availability of informative representations that encode the elementary geometric and semantic layout of building interiors, i.e., bounding walls, floors and ceilings, as well as the presence of portals like doors and the location of furniture. This information allows, among other tasks, to provide an informative visualization of the environment, organize its space and plan (respectively, simulate) the navigation of humans or robots inside it.

The described architectural layout is often summarized into {as-built} 2D or 3D floor plans, which describe an environment in a structured, yet compact form. Being based on existing real-world geometry, as-built plans must be derived from measured data.
Practitioners in architecture and engineering often resort to labor-intensive and tedious manual pipelines, in which an artist manually creates the desired model using interactive modeling tools (e.g., Trimble SketchUp~\cite{Sketchup}), guided by a combination of measured dimensions, point clouds and pictures. 
To increase the efficiency of this process, researchers have investigated the problem of extracting structured models of interiors automatically from acquired data~\cite{Pintore:2020:SA3}. The proposed approaches can greatly speed up the process of generating indoor models; still, they rely on rich 3D or 2D input data whose acquisition is time-consuming and logistically complex, requiring extended data capture sessions by experienced users or even trained personnel, using costly devices like laser scanners.
This is certainly true for laser-scanned 3D point clouds, but also for \mbox{RGB(-D)} videos~\cite{Liu:2018:FAU} or sets of overlapping panoramic images~\cite{Bao:2014:U3L, Cabral:2014:PPC, Pintore:2019:AMC} that are used in a number of recent pipelines. Clearly, the high requirements in terms of data acquisition hinder the large-scale creation of models of existing interiors, which in turn creates a bottleneck for the adoption of new, disruptive workflows for their management.

This bottleneck forms the starting point of our work. We explore the potential role of a much more modest data source that can be captured effortlessly, namely the \emph{walk trajectory of a person inside the space}.
Thanks to recent advances in data-driven inertial navigation~\cite{Yan:2018:RRI, Yan:2019:RRN, Chen:2019:DNN}, such trajectories can conveniently be recovered from the sensor readings of a consumer-level smartphone carried in a bag or pocket while walking, and without the need to actively operate the device (e.g., to take pictures). Using trajectories potentially turns every person into a passive {\em citizen sensor} that gathers data for the extraction of floor plans, and can pave the way to systematically mapping the world's interior spaces.

At first glance, deriving semantically enriched 2D floor plans from simple walk trajectories may seem an impossible and outrageous endeavour.
Indeed, such minimalistic input makes the task challenging and fundamentally ill-posed: a trajectory only provides incomplete evidence of which locations are {\em not} occupied by walls or furniture, but gives no direct evidence for their locations.
However, we argue that interiors are designed and furnished to meet specific functional requirements connected to the way people live and work, and therefore also to their way of moving through them~\cite{gibson:79, Monszpart:2019:IIG}. Consequently, it may be possible to leverage modern (deep) machine learning technology to uncover the latent relation between peoples' movement patterns and the locations of architectural elements relevant for constructing a floor plan.

To achieve this goal,we propose a multi-stage pipeline, in which each stage extracts one of the three fundamental elements of a (semantically annotated) floor plan: (i)~the 2D footprint of the interior space; (ii)~the location of portals (i.e., doors) in the wall boundary; (iii)~the space of the interior footprint occupied by furniture. Each step is cast as a machine learning task and solved with an appropriate model specifically tailored for it.
The benefit of splitting the computation into individual steps is twofold: on the one hand, it reflects the fact that the extraction of some elements of the floor plan depends on the knowledge of others (e.g., detecting openings relies on knowing the locations of walls); on the other hand, it allows one to keep the complexity of the neural architectures low, such that they can be trained with limited amounts of annotated data.
Underlying to the entire pipeline is a unified representation of the elements that compose a floor plan (i.e., walls, doors, furniture) as a 2D grid of cells and edges. Depending on the specific task, this representation can readily be converted into a top-down (i.e., birds-eye) raster image (e.g., to predict occupancy information such as the space taken by furniture) or into a graph structure (e.g., to localize opening in closed wall boundary loops).

In summary, this paper makes two main contributions.
At the \emph{conceptual level} we show for the first time that the a priori expectations about building interiors are indeed sufficient to derive surprisingly rich and accurate 2D floor plans only from sparse observations of empty locations.
At the \emph{technical level}, we develop a practical, data-driven system to recover the main architectural elements from a walk trajectory traversing the free space.

\begin{figure*}
  \includegraphics[width=\textwidth]{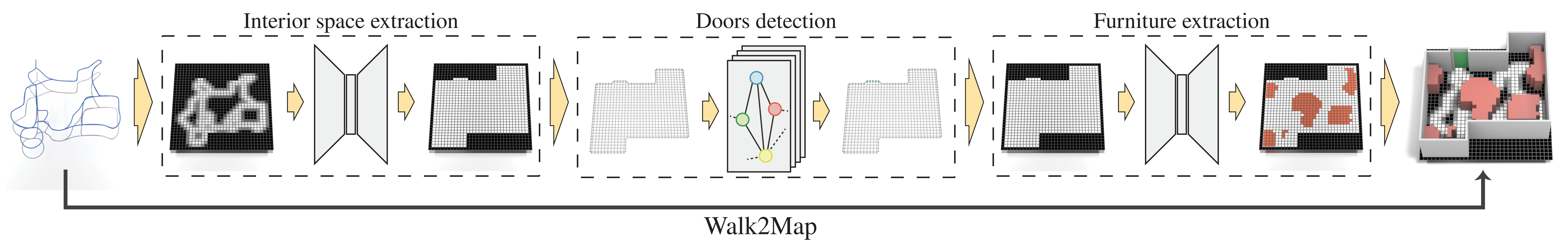}\vspace{-0.5em}
  \caption{Overview of our pipeline. Given an input walk trajectory, we project it onto the 2D domain of the floor plan and feed this representation in a cascaded manner to three neural networks. These extract, in order: the 2D footprint of the interior space; the location of doors on the wall boundary; the space occupied by furniture.}\vspace{-1em}
  \label{fig:pipeline_overview}
\end{figure*}

%-------------------------------------------------------------------------

%-------------------------------------------------------------------------

\section{Related Work}
\label{sec:related_work}

To the best of our knowledge, the extraction of floor plans from walk trajectories has not been studied before. Here, we analyze existing works about floor plan extraction, inertial odometry and deep learning for indoor scenes that are most closely related to our approach in terms of goals and technical formulations.

\paragraph*{Floor plan extraction from 2D/3D inputs.} 
The automatic extraction of interior models from scanned data has been studied in different fields of research over the last decades. The common goal of existing techniques is to compute a {\em structured} output model~\cite{Ikehata:2015:SIM, Pintore:2020:SA3} that describes the geometry of the main permanent entities while at the same time encoding the overall structure of the environment and the location of smaller elements (e.g., portals, furniture, fixtures).
In most pipelines, such a model takes the form of a {\em 2D floor plan}, a top-down view that encodes the 2D footprint of the interior space bounded by walls, and often also the location of the main objects in it. In some cases the 2D representation is extended into a fully-fledged 3D model.

Despite their different traits, existing approaches can be categorized~\cite{Pintore:2020:SA3} based on whether they start from 3D geometric observations (such as 3D point clouds), 2D visual data (e.g., RGB images or videos) or a mixture of the two.

Methods based on high-quality 3D inputs with little noise (e.g., laser-scanned point clouds) typically aim at modeling the geometry of walls and other permanent structures with the highest possible accuracy.
In most cases, the 3D input is projected onto the 2D ground plane and partitioned into a set of convex 2D polygons, either by triangulation~\cite{Turner:2012:WBA} or by fitting 2D lines and intersecting them~\cite{Oesau:2014:ISR, Mura:2014:ARD, Ochmann:2016:ARP}. Floor plans are then computed as the union of polygons that fall inside the walls, using techniques like binary min-cut~\cite{Oesau:2014:ISR}, $k$-medoids clustering~\cite{Mura:2014:ARD, Ikehata:2015:SIM} or multi-label optimization~\cite{Ochmann:2016:ARP, Mura:2016:PRM}. 
More recently, researchers have proposed using neural networks to extract the 2D coarse room shapes and converting these into vectorized floor plans by solving a sequence of shortest path problems~\cite{Chen:2019:FIC}.
Several methods go one step further and either lift the floor plans to 3D by extruding them along the vertical axis~\cite{Mura:2014:ARD, Ochmann:2016:ARP}, or perform the space partitioning with 3D polyhedra rather than 2D polygons~\cite{Oesau:2014:ISR, Mura:2016:PRM, Ochmann:2019:ARF}. 
Reconstruction from dense scan data is fairly robust and mature. The price to pay is that the approach does not scale well, due to the need to cover the scene (almost) completely with high-quality scans via complex data capture workflows, often with expensive equipment operated by trained personnel. 

A second category of approaches avoids the need for 3D scans and instead uses RGB or black and white images, which can be acquired with consumer cameras. Multi-View Stereo (MVS) has often been used in this context, for instance using the assumption of mutually orthogonal walls (i.e., {\em Manhattan-World})~\cite{Furukawa:2009:RBI} or geometric reasoning~\cite{Cabral:2014:PPC, Pintore:2018:3FP} to cope with the lack of texture typical of indoor scenes.
A major disadvantage of these approaches is that they require a high number of input images with high overlap for MVS to yield sufficient 3D evidence.
Other researchers have therefore focused on reducing the number of input images needed. Under very strong assumptions about the scene (e.g., a box-like layout~\cite{Hedau:2009:RSL} or lack of clutter~\cite{Lee:2009:GRS}), even a single image with limited field-of-view can be used.
More recently, single-image approaches~\cite{Zhang:PAN:2014, Yang:2016:E3R, Xu:2017:P2C}, including some based on deep learning~\cite{Sun:2019:HLR, Yang:2019:DDN, Nie:2020:TJL}, handle more general scenes and achieve better robustness using single images with large field of view -- often panoramas. 
Compared to dense 3D inputs, 2D images can be captured by less trained people and with consumer devices. However, single images with limited field of view are in practice not enough to capture realistic interiors, while panorama photography again requires a planned and dedicated recording effort.

Similar limitations affect hybrid methods that rely on streams of \mbox{RGB-D} data as input~\cite{Liu:2018:FAU}: although fairly reliable reconstruction algorithms are available~\cite{Zollhofer:2018:SA3}, capturing the raw RGB-D data in a way that enables a complete reconstruction of the environment is a non-trivial and labor-intensive exercise.

Our work differs in spirit from all mentioned approaches. We explore to what degree floor-plan reconstruction is still possible from input data that is much less expressive, but a lot easier to record -- namely a curve describing a walking trajectory through the indoor space. That data can be obtained with virtually no effort and without active human engagement, for instance using modern data-driven inertial odometry. This way, the reconstruction can scale well to systematic, large-scale mapping of interiors.

\paragraph*{Data-driven inertial odometry.}
Motion tracking in GPS-denied interiors has been an active research topic in the last decades. If image data captured by a moving camera are available, Visual Odometry~\cite{Nister:2004:VOD, Scaramuzza:2011:VOT} provides a solution, by localizing each camera pose relative to the previous one(s). This approach, nowadays mature on its own, can be further improved by Visual-Inertial Odometry~\cite{Leutenegger:2015:KVI}, which incorporates rotational velocities and linear accelerations measured by an Inertial Measurement Unit (IMU) in the tracking.
Despite their robustness, both purely visual and visual-inertial odometry require a continuously operating camera during the movement, which is impractical in many scenarios. 

In principle, the output of the IMU alone is sufficient to recover full location information by double-integration. In practice, however, standard IMUs provide noisy measurements, which rapidly accumulate during the integration process and make the resulting positions unusable. IMUs accurate enough for long-term operation have been available for some time, but are restricted to high-end scenarios (e.g., avionics) due to their prohibitive cost.

In the last few years, researchers have explored data-driven approaches to compensate for the drift inherent to double-integration. The basic idea is to train a model to regress a velocity relative to the previous position based on a sequence of IMU readings; the resulting velocity can then be used in different ways to compute actual positions. 
Yan et al.~\shortcite{Yan:2018:RRI} use the velocity regressed by using Support Vector Machines (SVM) to model low-frequency corrections for the raw acceleration values and use them to adjust the results of double-integration. Chen and colleagues~\shortcite{Chen:2019:DNN} employ a Recurrent Neural Network (RNN) to regress displacements that are used directly to compute the new positions.
The recent creation of large-scale datasets for inertial odometry~\cite{Yan:2019:RRN, Chen:2020:DLP} has made it possible to refine these techniques, and further improvements can be expected in the near future.

In our work, the trajectories generated with data-driven inertial odometry represent input data, from which we extract semantically rich 2D floor plans.

\paragraph*{Deep learning for indoor scene synthesis.}
Given a description of the architecture of an indoor space, a recurring question in interior design is how to find a furniture layout that satisfies the physical constraints of the space while being semantically practical (in the case of real spaces) or plausible (in the case of virtual spaces). 

The placement of furniture objects has been traditionally done using various hand-crafted priors~\cite{Merrell:2011:IFL, Yeh:2012:SOW, Yu:2011:MHA}. More recently, a number of data-driven approaches have been proposed~\cite{Fisher:2012:ES3, Ma:16:A3I, Savva:16:PLI, Fu:2017:ASI, Qi:2018:HIS, Wang:2018:DCP, Li:2019:GGR, Ritchie:2019:FFI}. 
In general, these techniques can be divided into two main groups~\cite{Wang:2019:PPI}: {\em object-oriented} methods~\cite{Fisher:2012:ES3, Qi:2018:HIS, Li:2019:GGR}, which explicitly model the objects in the scene and their properties (e.g., type, relative position), and {\em space-oriented} approaches~\cite{Wang:2018:DCP, Ritchie:2019:FFI}, which instead adopt a representation of the scene as a set of discrete spatial regions (i.e., a regular grid) and determine which objects fall in which region. The latter approaches work by predicting top-down view images of the scene and are thus largely based on Convolutional Neural Networks (CNNs).
The two strategies are combined by Wang et al.~\cite{Wang:2019:PPI} in their recent work, in which the scene layout is planned with an object-oriented, graph-based generative network, while the instantiation of objects relies on an image-based network.

These approaches differ conceptually from our work, as we aim to {\em reconstruct} existing scenes rather than {\em synthesize} new ones. Nevertheless, we borrow a number of technical ideas from scene synthesis, including the representation of the spatial domain as a top-down image~\cite{Wang:2018:DCP} and the use of convolutions on suitably constructed graphs~\cite{Wang:2019:PPI} to locate openings in the wall boundary of the scene. 

%-------------------------------------------------------------------------

%-------------------------------------------------------------------------

\section{Method and data overview}
\label{sec:method}

Our goal is to map a walk trajectory through an indoor space to a map of its walls, doors and furniture objects -- the elements most commonly included in architectural floor plans~\cite{Liu:2017:RRF}. We specifically target single-room, \emph{Manhattan World} environments (i.e., with mutually orthogonal walls, floor and ceiling), as these represent the large majority of real-world cases. 

We use three staged neural networks (see Fig.~\ref{fig:pipeline_overview}) to extract the three types of elements. By splitting the mapping task into sub-tasks, we can adapt the complexity of the networks to the specific learning problems: we use convolutional networks over a regular floor plan grid to learn occupancy information, that is, the footprint of the room (Sec.~\ref{sec:footprint_from_trajectory}) and the location of furniture (Sec.~\ref{sec:computing_furniture}), and a graph convolutional architecture to detect portals (Sec.~\ref{sec:locating_doors}) in walls.
Such a multi-stage pipeline also reflects the fact that each step depends on the previous one: locating openings in the wall boundary requires a notion of where the walls that delimit the room are, and furniture placement is influenced by the location of doors (e.g., furniture should not be placed in front of a door).

Each step is trained separately, with suitably adapted versions of the training data. 
At inference time, we apply the networks as a cascade, where the first step maps the input trajectory to a room footprint, the second step is fed both the trajectory and the footprint and detects the doors, and the third step receives the trajectory, footprint and doors as input and predicts the furniture. A post-processing step is applied after each step to regularize the raw predictions.

\subsection{Discrete floor plan representation} 

A floor plan can naturally be represented as a grid of discrete and uniformly-sized area units that cover the ground plane, i.e., a top-down (i.e., birds-eye) view with discrete ``pixels''.
Treating the floor plan as an image is convenient for space-oriented tasks that involve determining the occupancy of different locations; in particular, it makes the problem amenable to discrete 2D convolution~\cite{Wang:2018:DCP, Wang:2019:PPI}.

% Figure
\begin{figure}[!b]
  \includegraphics[width=\columnwidth]{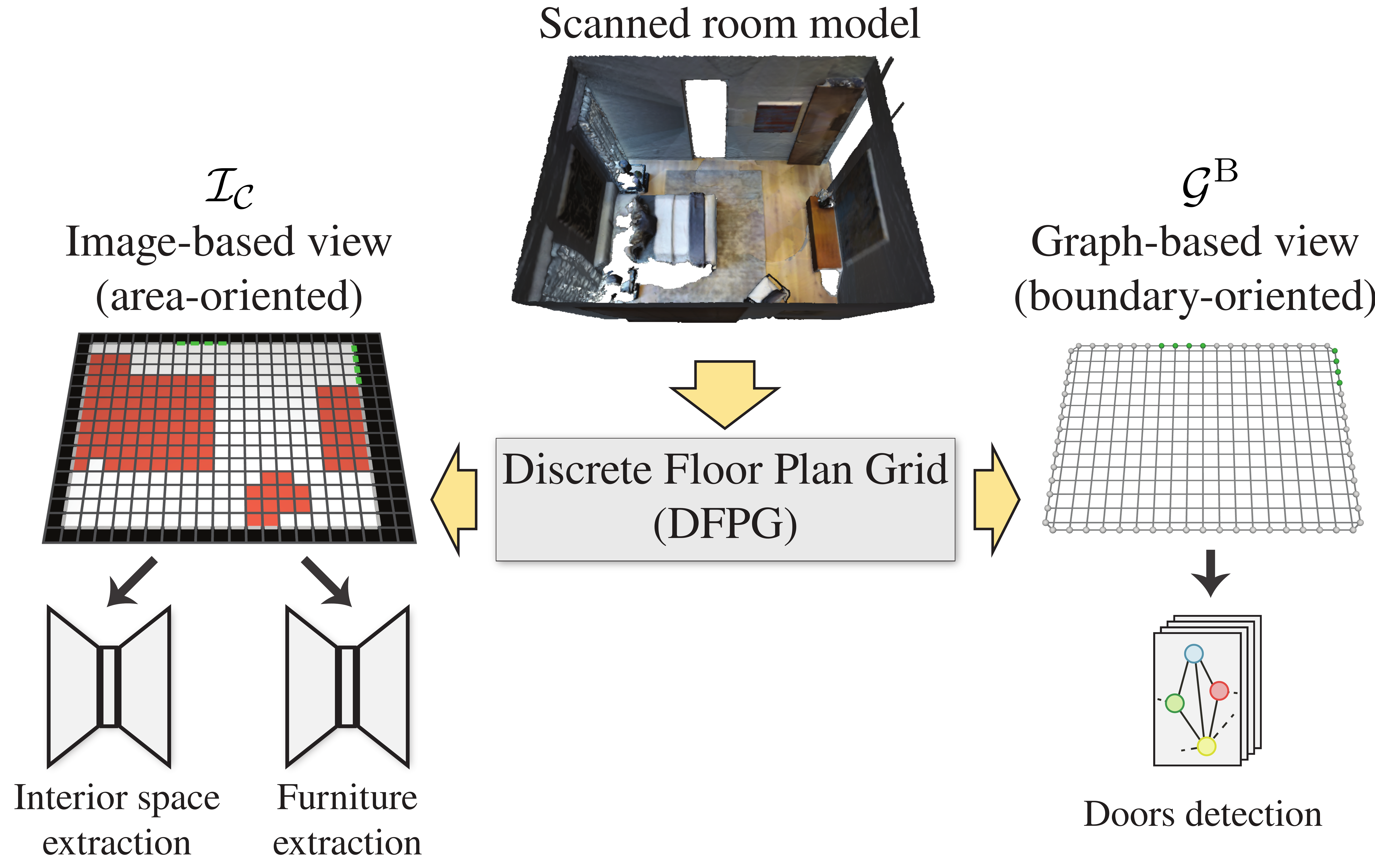}
  \caption{The two complementary views of an indoor model in our pipeline. An indoor space is represented as a {\em Discrete Floor Plan Grid (DFPG)}, from which one can obtain an image-based representation $\mathcal{I}_\mathcal{C}$ (for area-oriented learning tasks) as well as a graph-based representation $\mathcal{G}^{\mathrm{B}}$ (for boundary-oriented portal detection).}
  \label{fig:generated_floorplan_grid}
\end{figure}

We argue that some elements of an indoor scene -- specifically, walls, openings and wall-mounted flat objects -- can be better captured by a {\em boundary-oriented} representation. For this purpose, the thin, area-less interfaces between area units should be modeled explicitly, either densely as pixels of another image, or sparsely as nodes of a graph.

To efficiently switch between these alternative representations we internally store each floor plan as a mesh-based grid structure (see Fig.~\ref{fig:generated_floorplan_grid}), denoted as {\em discrete floor plan grid} or {\em DFPG} throughout the paper.

We explicitly model the set of cells $\mathcal{C} = \{ \mathbf{c}_1, \ldots, \mathbf{c}_{n_c}\}$ and the set of line segments at the cell boundaries, distinguishing between horizontal segments ${\mathcal{S}^{h} = \{ \mathbf{s}^h_1, \ldots, \mathbf{s}^h_{n_h}\}}$ and vertical segments ${\mathcal{S}^{v} = \{ \mathbf{s}^{v}_1, \ldots, \mathbf{s}^{v}_{n_v}\}}$. The cells $\mathcal{C}$ are arranged in a square grid of size $n \times n = n_c$, forming an image $\mathcal{I}_\mathcal{C}$.
Similar considerations hold for the sets $\mathcal{S}^{h}$ and $\mathcal{S}^{v}$, whose elements can be represented, respectively, as two images $\mathcal{I}_{\mathcal{S}^{h}}$, $\mathcal{I}_{\mathcal{S}^{v}}$ of size $(n+1) \times n = n_h$ and $n \times (n+1) = n_v$.

Both cells and segments carry labels that define their semantic attributes. The label $l_i$ of a cell $\mathbf{c}_i$ can take on the values $\{\mathbf{out}, \mathbf{in}, \mathbf{furn}\}$, which indicate that $\mathbf{c}_i$ is outside, inside or occupied by furniture, respectively. 
A horizontal or vertical segment is associated to one of the label values $\{\mathbf{door}, \mathbf{wall}, \mathbf{none}\}$, based on whether the segment is part of a door, a ``plain'' wall section or does not lie at the border between interior and outer space.
We denote by $\mathcal{I}^{\mathrm{sem}}_\mathcal{C}$, $\mathcal{I}^{\mathrm{sem}}_{\mathcal{S}^{h}}$, $\mathcal{I}^{\mathrm{sem}}_{\mathcal{S}^{v}}$ the {\em semantic maps} (label images) associated to $\mathcal{C}$, $\mathcal{S}^{h}$ and $\mathcal{S}^{v}$.
Finally, we define three more maps for $\mathcal{C}$: $\mathcal{I}^{\mathrm{in}}_\mathcal{C}$, a map of the interior space, obtained from $\mathcal{I}^{\mathrm{sem}}_\mathcal{C}$ by replacing label $\mathbf{furn}$ with $\mathbf{in}$; $\mathcal{I}^{\mathrm{free}}_\mathcal{C}$, a map of the {\em free} interior space obtained by replacing $\mathbf{furn}$ with $\mathbf{out}$; $\mathcal{I}^{\mathrm{walk}}_\mathcal{C}$, the raster representation of the input where each pixel stores the {\em inverse distance} to the walk trajectory.

\subsection{Training and test data} 

Since our model is based on machine learning, data is a crucial aspect of our work. We describe the datasets used for the different steps in the following, also providing an overview in Tab.~\ref{tab:sample_sets}.

\paragraph*{Synthetic data. } To date, there exists no open research dataset of real, measured indoor walk trajectories paired with ground-truth floor plans of the corresponding interiors. Hence, we created synthetic training examples from the Matterport3D dataset~\cite{Chang:2017:M3D}, which consists of annotated scanned models of $90$ \emph{real} indoor environments. 
We used the given annotations as well as newly generated ones to convert each valid room of each environment into a DFPG with semantic labels; details are given in Sec.~\ref{sec:implementation_details}. 
Of the \mbox{DFPGs} generated, we selected all those that include at least one door and one furniture element, and split them into mutually exclusive sets for training ($80\%$ of the data), validation ($10\%$) and testing ($10\%$). 
We created separate versions of the training, validation and test sets for each step of the pipeline (see Tab.~\ref{tab:sample_sets}), in image or graph representation as needed and using only the labels relevant for the respective steps.

For each DFPG generated we simulated a walk trajectory traversing the free interior space (the ground truth map $\mathcal{I}^{\mathrm{free}}_\mathcal{C}$), following recent work on indoor scene navigation~\cite{Kirsanov:2019:DDI}. The details of the simulation scheme are explained in Sec.~\ref{sec:implementation_details}. To better match the trajectories to our scene representation we convert them to raster images with the inverse distance transform, with a cut-off value of $0.5m$.

The interiors extracted from Matterport3D exhibit a variety of footprints and non-trivial arrangements of furniture and wall openings, but are limited in number ($\approx500$ samples). To extend the training set for the footprint extraction network (stage~1), we generated additional training samples ($50K$) from the LIFULL HOME'S dataset~\cite{Kiyota:2018:POI} of residential interiors, rasterized with the method of~\cite{Liu:2017:RRF}. This allowed us to adopt a curriculum learning strategy~\cite{Bengio:2009:CLe}, as explained in Sec.~\ref{sec:footprint_from_trajectory}.

\paragraph*{Real-world data. } In addition to the synthetic data derived from Matterport3D, we acquired a smaller test set composed of $20$ real-world trajectories through offices and private homes. To this purpose, we used the app developed by Yan et al.~\cite{Yan:2018:RRI} for Google Tango~\cite{Marder-Eppstein:2016:PT}, which was used in their inertial odometry work to acquire the ground truth data. 
To generate the corresponding ground truth floor plans we acquired 3D scanned models (using either a laser scanner or a Tango-enabled device) of the same environments in which the trajectories were acquired and annotated them manually.

\begin{table*}
\centering
\begin{tabular}{| c || c | c || c | c | c || c | c | c || c | c | c || c |}
\hline
\multirow{3}{*}{Source} & \multirow{3}{*}{Avg. area} & \multirow{3}{*}{Avg. diag.} & \multicolumn{10}{c|}{N. samples} \\
\cline{4-13}
& & & \multicolumn{3}{c||}{Stage~1} & \multicolumn{3}{c||}{Stage~2} & \multicolumn{3}{c||}{Stage~3} & Full pipeline \\
\cline{4-13}
 & & & train & valid. & test & train & valid. & test & train & valid. & test & test\\
\hline\hline
Matterport3D & 14.2$m^{2}$ & 6.1$m$ & 706 & 78 & 51 & 412 & 45 & 51 & 412 & 45 & 51 & 51 \\
\hline
LIFULL & 9.5$m^{2}$ & 4.3$m$ & 47500 & 2500 & - & - & - & - & - & - & - & - \\
\hline
Real-world & 21.3$m^{2}$ & 8.2$m$ & - & - & - & - & - & - & - & - & - & 20 \\
\hline
\end{tabular}
\caption{Statistics on the datasets used. For each data source we show the average area and diagonal of the environments and the size of the training/validation/test sets available at each stage of the pipeline, as well as the size of the test set used to evaluate the full pipeline. Note that LIFULL is used only to train the first network (see Sec.~\ref{sec:footprint_from_trajectory}) and that the real-world data is used only for end-to-end testing.}\vspace{-1em}
\label{tab:sample_sets}
\end{table*}

%-------------------------------------------------------------------------

%-------------------------------------------------------------------------

\section{Learning to extract floor plan elements}
\label{sec:learning}

This section details and discusses the neural architectures used in our pipeline. Implementation details are collected in Sec.~\ref{sec:implementation_details}. 

% ---------------------------------------------------------
% ---------      STEP 1: Interiors footprint      ---------
% ---------------------------------------------------------

\subsection{Interior space footprint from walk trajectory}
\label{sec:footprint_from_trajectory}

The most basic part of a floor plan, and the first recovered in our pipeline, is the 2D footprint of the interior space as enclosed by the bounding walls, that is, a binary segmentation of the 2D top-down view $\mathcal{I}_\mathcal{C}$ in which each raster cell is labeled as $0$ (outside the wall boundary) or $1$ (inside).
This information is recovered using as only input the inverse distance transform of the input trajectory, $\mathcal{I}^{\mathrm{walk}}_\mathcal{C}$.
Note that, since input and output are images over the same spatial domain, this process can be interpreted as an image-to-image translation~\cite{Isola:2017:ITC}. In our case, with limited training data, we abstain from using recently popular GANs and instead employ a conventional Encoder-Decoder network based on U-Net~\cite{Ronneberger:2015:UCN}.
This fully-convolutional architecture consists of a contraction path composed of several levels of convolution and pooling, and a symmetric expansion path with up-sampling (``transposed convolution") and regular convolutions.
The contracting path progressively aggregates low-level features into more abstract ones over larger receptive fields, while the expanding path super-resolves those features back to the pixel level, providing a dense segmentation at the original resolution.

Since only the footprint is relevant in this step, we use all available training samples from Matterport3D, including those that do not contain doors or furniture,
and used $50K$ additional samples extracted from LIFULL to support the training. Since these have rather simplistic room shapes and generally lower variability, we inject them via curriculum learning~\cite{Bengio:2009:CLe}: we first ``pre-train'' the network weights with the simpler, yet more abundant LIFULL samples; then, we complete the training with the more informative Matterport3D samples.

The predicted 2D segmentation maps $\mathcal{I}^{\mathrm{in}}_\mathcal{C}$ are integrated into the DFPG and explicit wall boundaries are extracted at the label transitions. As the labeling does not guarantee a closed, connected wall loop, gaps are repaired with a simple optimization scheme (Sec.~\ref{sec:implementation_details}).

% ---------------------------------------------------------

% ---------------------------------------------------------
% ---------       STEP 2: Locating doors       ---------
% ---------------------------------------------------------

\subsection{Locating doors on wall boundaries}
\label{sec:locating_doors}

Finding the location of doors in the walls is again formulated as a labeling problem, this time over the line segments $\mathcal{S}^{h}$ and $\mathcal{S}^{v}$ of the DFPG.
In principle one could use a 2D fully-convolutional network as in Sec.~\ref{sec:footprint_from_trajectory}, stacking the two channels $\mathcal{I}_{\mathcal{S}^{h}}$ for the horizontal and $\mathcal{I}_{\mathcal{S}^{v}}$ for the vertical edges and translating them to a two-channel segmentation map. 
However, this leads to an unnecessarily complex network over all line segments, including those not on walls.

Instead, we exploit the fact that the wall boundary of an interior forms a closed, orientable loop $\mathcal{S}^{\mathrm{B}} = \{ \mathbf{s}^{\mathrm{B}}_{1}, \ldots, \mathbf{s}^{\mathrm{B}}_{n_{\mathrm{B}}} \}$, where each segment $\mathbf{s}^{\mathrm{B}}_{i}$ has exactly one predecessor $\mathsf{prev}(\mathbf{s}^{\mathrm{B}}_{i})$ and one successor $\mathsf{next}(\mathbf{s}^{\mathrm{B}}_{i})$.
Without loss of generality, we orient $\mathcal{S}^{\mathrm{B}}$ in clock-wise order, so that the interior space is on the right side of a segment.
The boundary loop forms an undirected graph, with a node per line segment and an edge between adjacent segments. With a slight abuse of notation we refer to the $j$-th \emph{node} via the corresponding line segment $\mathbf{s}^{\mathrm{B}}_{j}$. 
Additionally, we connect each node to the line segment opposite to it in the wall boundary loop, as in PlanIT~\cite{Wang:2019:PPI}, to provide each node with more context of the room geometry.
In fact our graph is equivalent to the {\em initial} room representation of PlanIT; however, in that work the room graph is expanded at inference time with new nodes for {\em synthesized} openings and furniture, whereas we keep the set of nodes fixed and classify them as either a door or a plain wall section.

To inject global as well as local information about the environment, each node is associated with a feature vector obtained by concatenating the following attributes: 
\begin{itemize}
    \item the distance along the loop to the closest corner;
    \item the orientation (horizontal/vertical) of the wall segment;
    \item the inverse distance to the input trajectory for the $k=10$ cells of $\mathcal{I}_\mathcal{C}$ that lie to the right of the node;
    \item the 2D coordinates of the line segment in the DFPG.
\end{itemize}
This information is combined and propagated based on the graph connectivity using a Graph Convolutional Network (GCN) to obtain a binary labeling of the nodes as either a door ($\mathbf{door}$) or a plain wall section with no opening ($\mathbf{wall}$). 

A number of different models have been proposed to adapt traditional learning approaches to graph-structured data~\cite{Hamilton:2017:RLG}. 
We base our architecture on the Dynamic Edge-Conditioned GCN (DECGCN) by Simonovsky and Komodakis~\shortcite{Simonovsky:2017:DEF}. In that model, the features of adjacent nodes are combined with dynamic weights that are generated based on input edge features by trained \emph{filter-generating networks}.
To this end we define the following two edge features: 
\begin{itemize}
    \item the orientation dissimilarity $(1 - \mathbf{s}^{\mathrm{B}}_{i} \cdot \mathbf{s}^{\mathrm{B}}_{j})$ between the the two adjacent nodes (i.e., line segments);
    \item the $L_1$ distance $|\mathbf{s}^{\mathrm{B}}_{i}- \mathbf{s}^{\mathrm{B}}_{j}|_1$ between them in the DFPG.
\end{itemize}
In preliminary experiments comparing to the GCN architecture of~\cite{Kipf:2017:SSC}, we found edge-conditioned convolutions to be crucial for obtaining good classification results: with GCN without edge conditioning, there was a strong bias towards symmetric doors on opposite walls, whereas simply dropping the additional edges between opposite walls led to significantly lower classification performance.

After classification the node labels of the graph are again transferred to corresponding segment labels in the DFPG, where each segment is now assigned a label $\mathbf{wall}$, $\mathbf{door}$ or $\mathbf{none}$. 

% ---------------------------------------------------------

% ---------------------------------------------------------
% ---------           STEP 3: Furniture           ---------
% ---------------------------------------------------------

\subsection{Computing furniture location}
\label{sec:computing_furniture}

The location of doors has a major influence on the furniture placement in indoor spaces, as furniture should not obstruct natural passages. This is the intuition behind the separate detection of furniture as a last step.
Like the extraction of the interior space (Sec.~\ref{sec:footprint_from_trajectory}), this is a space-oriented, binary classification task, where the raster cells of the DFPG that lie inside the walls are labeled as either occupied by furniture ($\mathbf{furn}$) or part of the free space ($\mathbf{in}$).

We employ the same architecture as in Sec.~\ref{sec:footprint_from_trajectory}, with two small, but important differences.
First, we mask the (known) cells outside the walls (i.e., labeled $\mathbf{out}$ in the first step) and do not consider them in the loss function.
Second, we extend the input features: besides the inverse distance  $\mathcal{I}^{\mathrm{walk}}_\mathcal{C}$ to the walk trajectory we also include the binary map of the interior space, as well as the location of the doors in the form of the two segment label maps $\mathcal{I}^{\mathrm{sem}}_{\mathcal{S}^{h}}$ and $\mathcal{I}^{\mathrm{sem}}_{\mathcal{S}^{v}}$.
The input maps are arranged into a single four-channel image and fed to the network, which outputs a single-channel binary map with labels $\mathbf{furn}$ or $\mathbf{in}$.

% ---------------------------------------------------------

%-------------------------------------------------------------------------

%-------------------------------------------------------------------------

\section{Implementation details}
\label{sec:implementation_details}

\paragraph*{Neural network architectures. } The encoder-decoder networks used to find footprints (Sec.~\ref{sec:footprint_from_trajectory}) and furniture (Sec.~\ref{sec:computing_furniture}) are based on the same U-Net architecture described in the original paper~\cite{Ronneberger:2015:UCN}, with $3$ levels and skip connections between corresponding levels of the contracting and expanding paths; compared to the original configuration, we increased the number of features (activation maps) in the first convolution layer from $64$ to $128$.
We used the cross-entropy loss and trained the networks with ADAM~\cite{Kingma:2014:AAM}.
The networks are implemented in PyTorch~\cite{Paszke:2019:PIS}.

As mentioned in Sec.~\ref{sec:footprint_from_trajectory}, we first train the footprint network on the LIFULL dataset for $50$ epochs with a base learning rate of $0.005$ and decay rates $\beta_1=0.5$, $\beta_2=0.999$, switch to Matterport3D and train for another $200$ epochs, with a learning rate of $0.0001$ and the same values of $\beta_1$ and $\beta_2$.
As final model we retain the epoch with the highest validation accuracy.
Only for Matterport3D we employ data augmentation by random rotations as well as horizontal and vertical flipping.

For the furniture network, we set a base learning rate of $0.001$ with decay rates $\beta_1=0.5$ $\beta_2=0.999$, and trained for $200$ epochs, again selecting the model with the highest validation accuracy.

The door detection network consists of $5$ blocks, each composed of a convolutional layer with Dynamic Edge-Conditioned Filters~\cite{Simonovsky:2017:DEF}, followed by Batch Normalization and a ReLU non-linearity.
The convolution layers of the $5$ blocks have, respectively, output depth $64$, $128$, $128$, $64$ and $2$ (for binary classification); each of them employs a Feature Generation Network consisting of $3$ fully connected layers interleaved by ReLUs, with the first $2$ layers having an output size of $16$ and $32$, respectively.
The graph network is implemented in PyTorch Geometric~\cite{Fey:2019:FGR}.

All networks were trained on a single NVIDIA RTX 2080 Ti. The training time for one epoch was between $1$ and $3$ seconds for the graph-based network, as well as for the U-Net when training on Matterport3D ($<1K$ samples). On LIFULL ($50K$ samples) one epoch took about $190$ seconds.

\paragraph*{Extraction of DFPG representations. } We created a DFPG for each room of the Matterport3D and LIFULL (after vectorization) datasets, discarding rooms with non-axis-aligned wall segments as well as overly small ones for which the smallest side of the  bounding box is $<3.0m$ for LIFULL, respectively $<1.0m$ for Matterport3D (so as not to discard too many samples).
The 2D polygon of each valid room was rasterized into a DFPG grid with a cell size of $0.25m$. For simplicity we limit the floor plan size to $64 \mathrm{x} 64$ and discard the few rooms that do not fit that size.

For the Matterport3D samples, we used both the given annotations and newly generated ones to assignn the semantic labels (see Sec.~\ref{sec:method}).
We manually re-annotated all doors as they are rather inaccurate in the original data.
To determine furniture occupancy, we projected 3D primitives with the following labels onto the DFPG grid: 
\emph{bed}, 
\emph{chair}, 
\emph{sofa}, 
\emph{stool}, 
\emph{table}, 
\emph{cabinet}, 
\emph{chest\_of\_drawers}, 
\emph{furniture}, 
\emph{cushion},
\emph{plant},
\emph{sink},
\emph{tv\_monitor},
\emph{bathtub},
\emph{shower},
\emph{counter},
\emph{gym\_equipment},
\emph{seating}. 
Cells intersected by at least one of these primitives were labeled as $\mathbf{furn}$.

\paragraph*{Simulation of walking trajectories. } Our  simulation follows the ideas of DISCOMAN~\cite{Kirsanov:2019:DDI}: a set of positions sampled in the free space are ordered by solving a Traveling Salesman Problem (TSP) and turned into a path by connecting subsequent positions with shortest paths through the free space.
For path sampling we shrink the free space by a buffer of 1 cell ($=0.25m$) along the walls,  since people tend to avoid walking so close to a wall.
We sample locations in the empty space by stratified sampling: we divide the domain of the empty space in cells of size $2m\times2m$ and draw a sample in each cell with a probability proportional to the distance to the occupied space (i.e., furniture and walls). 

To obtain more informative trajectories, we augment this basic strategy in two ways. 
First, we generate a path loop around each connected region of furniture cells by connecting $4$ points sampled in the empty space around its bounding box. We simply drop path loops for which a sampled point lies in occupied/outer space, which adds some randomness to the process. These loops account for the fact that, while moving in tight indoor spaces for extended periods, people inevitably tend to move along the few possible paths around large obstacles, leading to circular trajectories around large objects.
Second, we connect the trajectory to each door by shortest paths, accounting for the movement of people in and out of the room. These paths are computed on the \emph{whole} empty space $\mathcal{I}^{\mathrm{free}}_\mathcal{C}$, including the cells next to walls, accounting for the fact that exiting a room through a door must always be possible.

\begin{figure}[!t]
  \includegraphics[width=\columnwidth]{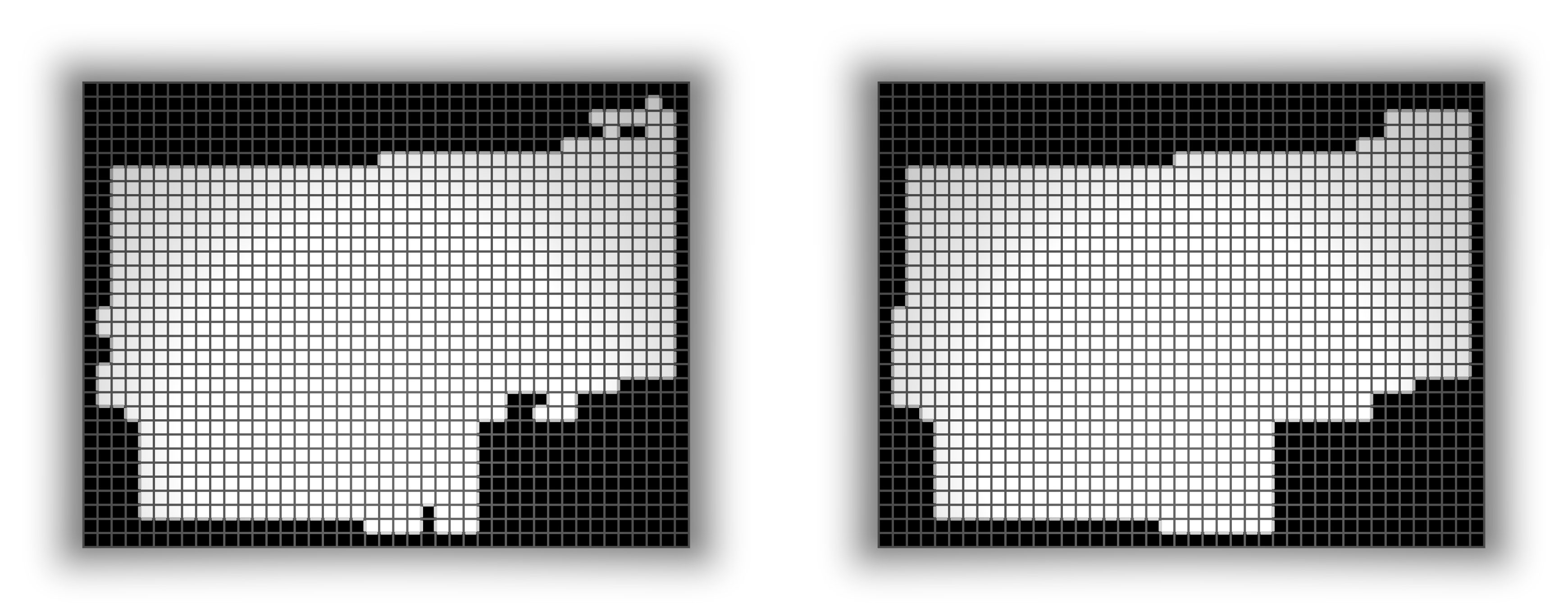}\vspace{-1em}
  \caption{Interiors footprint regularization. An energy-based optimization is applied to regularize the raw output (left) of the interior space prediction network, resulting in smoother boundaries that do not deviate from the predicted shape (right).}\vspace{-2em}
  \label{fig:footprint_regularization}
\end{figure}

\paragraph*{Postprocessing of raw predictions. } While the floor plan priors learned by the networks are surprisingly expressive, we have observed occasional mistakes with respect to basic low-level smoothness, including jagged boundaries of the interior space and isolated \textbf{door} or \textbf{furn} labels.
We therefore regularize the raw predictions by binary label swapping, according to simple continuity priors.

For both furniture and door prediction, we detect isolated cells (w.r.t.\ their 4-neighborhood, for furniture) or nodes (w.r.t.\ the previous and next segment, for doors) that have been classified with the target label and change their labels to $\mathbf{wall}$, respectively $\mathbf{in}$.
We additionally normalise the output of the doors prediction network by growing/shrinking them to a width of $4$ cells ($1$ m).

For interior footprints, we apply a more structured smoothing based on energy minimization, which corresponds to the binary labeling of a Markov Random Field associated to the predicted 2D maps $\mathcal{I}^{\mathrm{in}}_\mathcal{C}$. 
This is a well-established technique for image restoration and shape regularization~\cite{Li:1995:MRF}. 
We consider a data term (for unary potentials) that penalizes swapping the label $\mathbf{in}$ with $\mathbf{out}$ by a penalty $\gamma_{\, \mathrm{in} \rightarrow \mathrm{out}} = 4$ and the inverse swapping (i.e., assigning label $\mathbf{out}$ to a pixel predicted $\mathbf{in}$) by a penalty $\gamma_{\, \mathrm{out} \rightarrow \mathrm{in}} = 1$. The data term is paired with a smoothness term (for binary potentials corresponding to adjacent pixels) based on the Potts model, penalizing by $\gamma_{\, \mathrm{border}} = 2$ the assignment of different labels to adjacent locations.
Note that our choice of the data term (with $\gamma_{\, \mathrm{in} \rightarrow \mathrm{out}} > \gamma_{\, \mathrm{out} \rightarrow \mathrm{in}}$) penalizes {\em shrinking} the footprint rather than expanding it. 
For the sake of practicality, we minimize the resulting energy function via Integer Programming, using the GUROBI solver~\cite{Gurobi:2020:GOR}. Note, however, that the global optimum can be obtained more efficiently using graph cuts~\cite{Kolmogorov:2004:WEF}.

As shown in Fig.~\ref{fig:footprint_regularization}, this regularization step significantly improves the extracted room boundaries.

%-------------------------------------------------------------------------

%-------------------------------------------------------------------------

\section{Results}
\label{sec:results}

We present in this section the evaluation of our approach, using both simulated (Sec.~\ref{sec:eval_synthetic}) and measured (Sec.~\ref{sec:eval_real-world}) trajectories. 
We compare against a baseline derived from image-to-image translation~\cite{Isola:2017:ITC} in Sec.~\ref{sec:baseline_comparison}.

% ------------------------------------------------------
% ---------       Simulated trajectories       ---------
% ------------------------------------------------------

\subsection{Evaluation on simulated trajectories}\label{sec:eval_synthetic}

We use the test set extracted from  Matterport3D (see Sec.~\ref{sec:method}) for a quantitative as well as qualitative analysis of our approach. Training was performed with the same data characteristics, so this evaluation provides an assessment of our approach in the presence of a representative training set, unaffected by the independent issue of domain shift between the training and test distributions.

Tab.~\ref{tab:quantitative_results_all} shows our quantitative analysis, which we performed by running the complete cascade end-to-end from trajectories to full floor-plans, including the intermediate regularization steps. As an unavoidable side effect, each step inherits the errors of the previous one.
For each feature of the floor plan (interior space, doors, furniture), we report precision and recall values averaged over all test samples and the corresponding F1 score.
For both interior space extraction and furniture detection we consider a ground truth cell to be predicted correctly if there is at least one predicted cell with the same label at a distance $\leq 0.25m = 1~\textrm{cell}$.
Similarly, we define the distance between a true and a predicted door as the minimum Euclidean distance between any of their constituting line segments and declare the prediction correct if that distance is $\leq 0.25m = 1~\textrm{cell}$.
For comparison, the average diameter of a room is $6.1m \approx 24~\textrm{cells}$ for the Matterport3D samples and $8.2m \approx 33~\textrm{cells}$ for the real-world ones.
This way of evaluating the results is particularly important for doors, as their ground truth locations are defined with respect to the ground truth wall boundary, which will in general differ from the predicted one.

The statistics show that {\em Walk2Map} achieves excellent performance on the interior space extraction, with precision, recall and F1 values above $96\%$. The remaining two steps, i.e. door and furniture prediction, also attain high scores of over $70\%$ and almost $80\%$, respectively, despite the fact that they suffer from the errors accumulated in the cascade.
It is important to note that both the extraction of the interior space footprint and the prediction of furniture are very ambiguous. The much better performance on the former can be attributed on one hand to the larger available training set (by including LIFULL) and on the other hand to the higher degree of regularity, which constrains the wall layout more tightly, both implicitly in terms of the network weights and explicitly through regularisation.

Interestingly, the results in Tab.~\ref{tab:quantitative_results_all} also show that for the area-oriented networks (i.e., for interiors and furniture prediction) the recall values are better than precision, that is, there is a tendency towards false positives (``overshooting''), whereas for doors the opposite is true. We attribute this behavior of the door prediction network to the stark imbalance between the two labels ($\mathbf{door}$ and $\mathbf{wall}$) considered.

\begin{table}[t!]
\centering
\renewcommand{\arraystretch}{1.25}
\begin{tabular}{| c | l || c | c | c || c | c | c |}
\cline{3-8}
\multicolumn{2}{r||}{} & \multicolumn{3}{c||}{Walk2Map} & \multicolumn{3}{c|}{pix2pix} \\
\cline{2-8}
\multicolumn{1}{c|}{} & \multicolumn{1}{c||}{Element} & Prec. & Rec. & F1 & Prec. & Rec. & F1 \\
\hline\hline
\parbox[t]{2mm}{\multirow{3}{*}{\rotatebox[origin=c]{90}{Simulated}}} & Interior & $.96$ & $.99$ & $\mathbf{.97}$ & $.70$ & $.82$ & $.76$ \\  
\cline{2-8}
& Doors & $.77$ & $.72$ & $\mathbf{.74}$ & $.50$ & $.32$ & $.39$ \\  
\cline{2-8}
& Furniture & $.77$ & $.79$ & $\mathbf{.78}$ & $.62$ & $.39$ & $.48$ \\  
\hline\hline
\parbox[t]{2mm}{\multirow{3}{*}{\rotatebox[origin=c]{90}{Measured}}} & Interior & $.96$  & $.97$ & $\mathbf{.96}$ & $.74$ & $.71$ & $.72$ \\
\cline{2-8}
& Doors  & $.62$  & $.48$ & $\mathbf{.54}$ & $.18$ & $.20$ & $.19$ \\
\cline{2-8}
& Furniture  & $.85$  & $.60$ & $\mathbf{.70}$ & $.52$ & $.41$ & $.46$ \\
\hline
\end{tabular}
\caption{Quantitative evaluation. For each element of the floor plan (i.e., interior space, doors, furniture), we show the average precision (\emph{Prec.}) and recall values (\emph{Rec.}) as well as the F1 score (\emph{F1}). Rows \emph{Simulated} refer to the evaluation on simulated trajectories, rows \emph{Measured} to the one on real-world, measured trajectories. We show the results of our method (columns~\emph{Walk2Map}) alongside those of the baseline (\emph{pix2pix}) described in Sec.~\ref{sec:baseline_comparison}.}\vspace{-2em}
\label{tab:quantitative_results_all}
\end{table}

\begin{figure*}[t!]
  \centering
  \includegraphics[width=\textwidth]{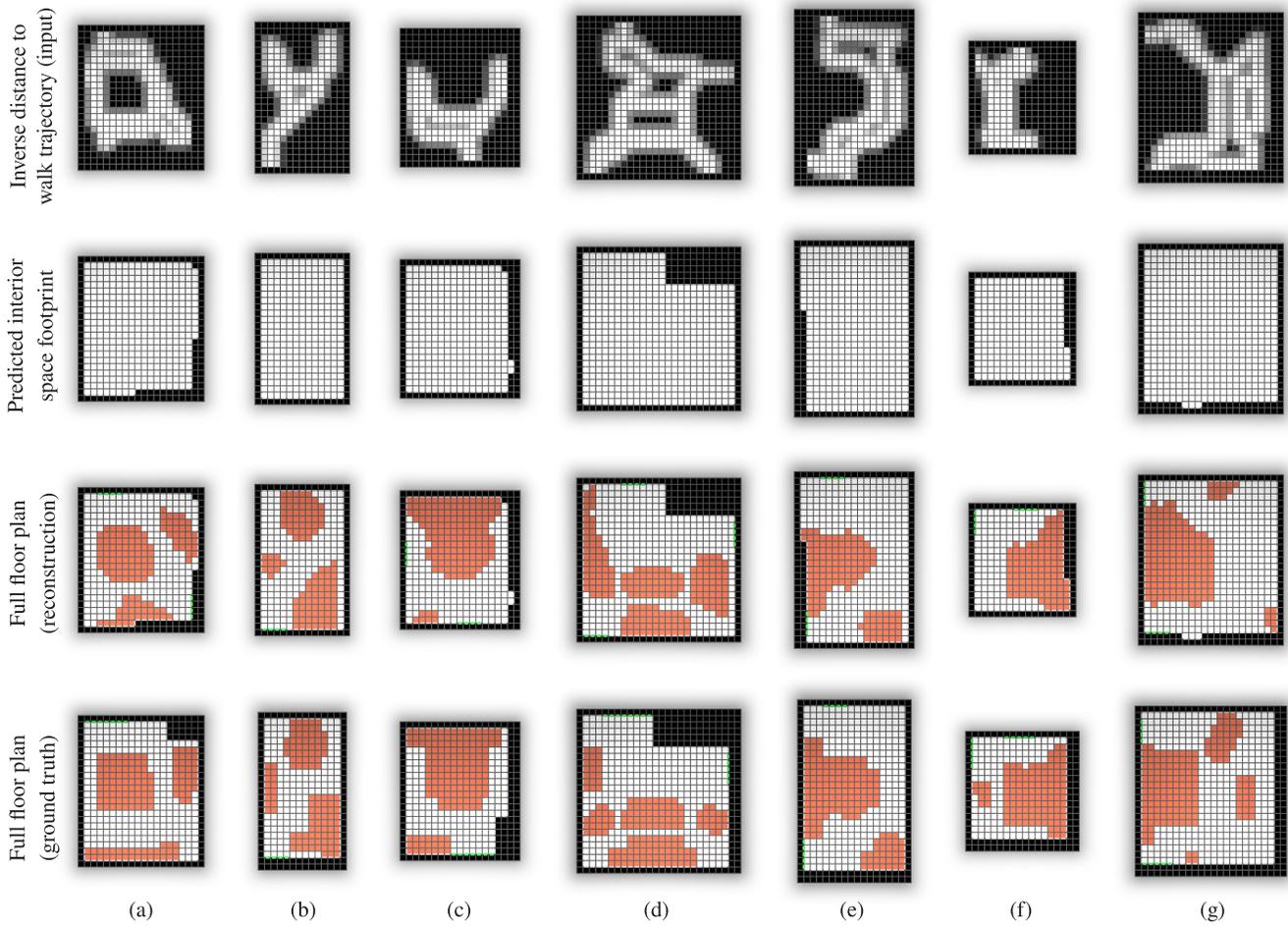}\vspace{-1em}
  \caption{Qualitative evaluation on simulated trajectories (2D visualization). From top to bottom: inverse distance to the trajectory that forms the input of our pipeline; predicted interior space footprint; full floor plan with predicted doors and furniture; ground-truth floor plan. The samples are randomly picked from the Matterport3D test set. See Fig.~\ref{fig:qualitative_evaluation_3D_models} for 3D visualizations.}\vspace{-0.5em}
  \label{fig:qualitative_evaluation_floorplans}
\end{figure*}

\begin{figure*}[!ht]
  \includegraphics[width=\textwidth]{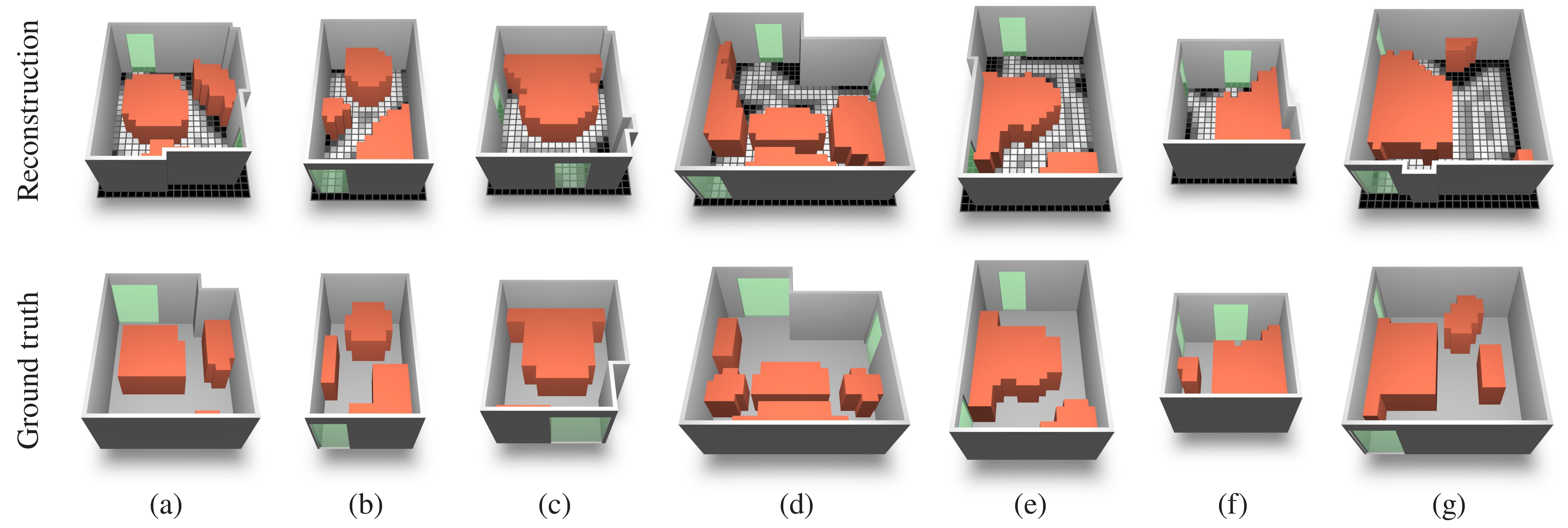}\vspace{-1em}
  \caption{Qualitative evaluation on simulated trajectories (3D visualization). We show the full floor plan models reconstructed with our method (top) and the corresponding ground truth (bottom). The input inverse distance field is overlaid on the ground plane of the reconstructed model. The same samples are shown in 2D in Fig.~\ref{fig:qualitative_evaluation_floorplans}.}
  \label{fig:qualitative_evaluation_3D_models}
\end{figure*}

\begin{figure*}[t!]
  \centering
  \includegraphics[width=\textwidth]{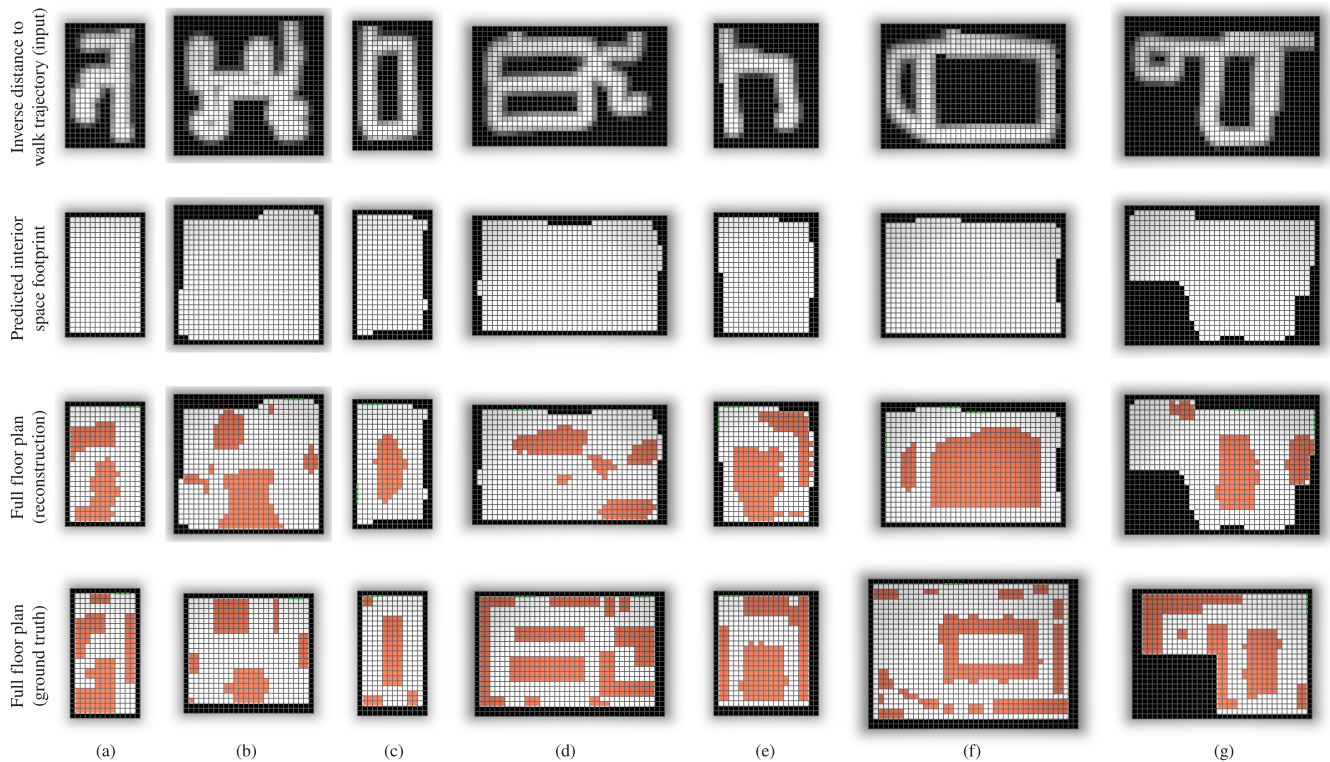}\vspace{-0.5em}
  \caption{Qualitative evaluation on real-world trajectories (2D visualization). The layout follows the one used in Fig.~\ref{fig:qualitative_evaluation_floorplans} for simulated trajectories. The samples are a random selection from the test set of real-world trajectories (Sec.~\ref{sec:method}) and are visualized in 3D in Fig.~\ref{fig:qualitative_evaluation_3D_models_rw}.}\vspace{-2em}
  \label{fig:qualitative_evaluation_floorplans_rw}
\end{figure*}

The overall effectiveness of {\em Walk2Map} is confirmed, and perhaps better demonstrated, by a visual analysis of the results.
A selection of reconstructed floor plans, shown in Fig.~\ref{fig:qualitative_evaluation_floorplans}, confirms that our pipeline actually succeeds in reconstructing largely correct interiors from the walking trajectory alone. Smaller concavities are sometimes missed (a, c), partly explaining the drop in precision. The main shape is in general reconstructed rather well (d).

Doors are also reliably detected, as best seen in the 3D visualizations of Fig.~\ref{fig:qualitative_evaluation_3D_models}. Errors do occur, but are nevertheless plausible in the context of the room and furniture.

The furniture detection network generally manages to successfully localize the main blocks of furniture; the relatively lower precision and recall compared to the other area-oriented task (interiors prediction) are largely due to inaccurate reconstructions of the ambiguous block boundaries. It does happen that entire furniture blocks are missed or incorrectly added when this is plausible given the trajectory (see Fig.~\ref{fig:failure_cases}).
We emphasize that furniture placement is an extremely difficult task, as the trajectory only provides local evidence for the \emph{absence} of furniture, whereas its presence must be inferred entirely from the prior and long-range context. 

\paragraph*{Graph-based vs image-based door detection network.} We used the Matterport3D samples to evaluate the performance of a door detection step based on a CNN architecture as an alternative to our graph-based model.
In particular, we tested the same U-Net architecture used for interior footprint prediction. 
The input to this network is a 4-channel image, composed of the two edge maps $\mathcal{I}_{\mathcal{S}^{h}}$, $\mathcal{I}_{\mathcal{S}^{v}}$, of the cells image $\mathcal{I}_\mathcal{C}$ and of the image $\mathcal{I}^{\mathrm{walk}}_\mathcal{C}$ with the inverse distance to the trajectory; the output is a 2-channel image containing the predicted edge maps encoding the position of the boundary walls, including the location of doors.
Note that, as done for the furniture detection network (see Sec.~\ref{sec:computing_furniture}), we masked the pixels that do not correspond to walls.

The evaluation yielded a high accuracy of $96\%$ but with an unacceptably low recall of $9\%$: the predicted door locations are generally correct but the number of actual doors missed by the network is so high to make its predictions unusable. 
Moreover, as already mentioned in Sec.~\ref{sec:locating_doors}, this architecture has a much higher number of parameters compared to its graph-based counterpart ($8.5M$ vs. $1.15M$), which makes its training significantly harder. In our tests, we had to reduce the batch size to $2$ to prevent the network from labeling the whole boundary loop as $\textbf{wall}$.

% -------------------------------------------------------
% ---------       Real-world trajectories       ---------
% -------------------------------------------------------

\subsection{Evaluation on real-world, measured trajectories}\label{sec:eval_real-world}

We also tested our pipeline -- trained on the simulated data -- on a set of $20$ real-world indoor trajectories (see Sec.~\ref{sec:method}). Since these samples come from a different generation process compared to the ones used in training, this evaluation is not fully indicative of the true capabilities of the pipeline; nevertheless, it allows us to conjecture about the effectiveness of \emph{Walk2Map} in real-world scenarios.

Tab.~\ref{tab:quantitative_results_all} (rows `Measured') shows the quantitative results obtained by running the real-world trajectories through the whole pipeline (without retraining). 
The statistics are computed following the same procedure presented in Sec.~\ref{sec:eval_synthetic}; since the measured trajectory is not registered with the 3D scan used to generate the ground-truth, we align the predicted and the ground-truth floor plans by overlapping the midpoint of their bounding boxes.
The statistics are generally consistent with those obtained for simulated data: the interior prediction network achieves very good results (both precision and recall $> 96\%$), whereas door and furniture prediction exhibit lower performance (with recall values of $48\%$ and $60\%$, respectively).
As discussed in Sec.~\ref{sec:eval_synthetic}, the lower scores on doors and furniture prediction are partly explained with the error accumulated from the previous steps. With specific reference to doors, however, we additionally observed slight and variable delays in the start and stop of the recordings of the trajectories; since doors typically correspond to the endpoints of a trajectory, such delays resulted in door signatures that are more irregular compared to the ones in the training samples, leading to an increased number of false negatives.

Comparing the rows `Simulated' and `Measured' in Tab.~\ref{tab:quantitative_results_all} clearly shows that the performance on real-world trajectories is lower than that on simulated data. This can be explained with the inevitable domain shift between simulated training data and real test data. \emph{Learning} to generate realistic synthetic trajectories so as to avoid this shift is an interesting avenue for future work.

\begin{figure*}[t!]
  \includegraphics[width=\textwidth]{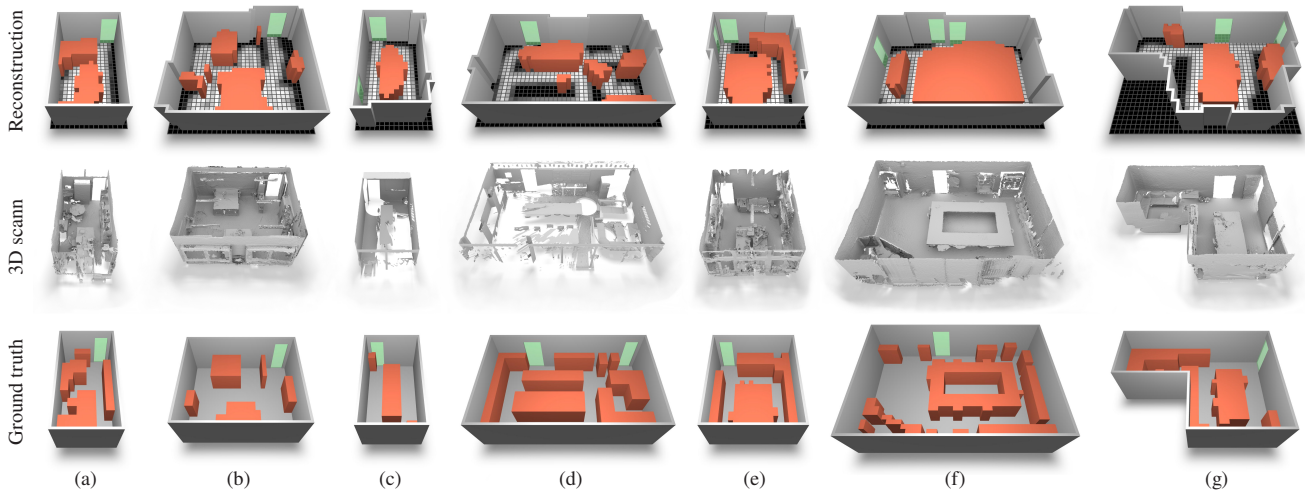}\vspace{-0.5em}
  \caption{Qualitative evaluation on real-world trajectories (3D visualization). From top to bottom: full floor plan models reconstructed with our method; 3D scans of the corresponding environment; ground-truth floor plan models, obtained by manually annotating the scans. The same samples are shown in a 2D top-down view in Fig.~\ref{fig:qualitative_evaluation_floorplans_rw}.}\vspace{-2em}
  \label{fig:qualitative_evaluation_3D_models_rw}
\end{figure*}

The visual analysis of the results (Figs.~\ref{fig:qualitative_evaluation_floorplans_rw} and~\ref{fig:qualitative_evaluation_3D_models_rw}) gives a more intuitive impression of the floor plans predicted from real trajectories. 
Compared to the simulated case, errors in the localization of furniture blocks are more common (d, g), and door localization suffers from more false positives (e, f, g) and false negatives (d). 
Nevertheless, the general structure of the environment is correctly recovered and its elements are arranged in a meaningful way. 

Overall, the experiments confirm that reconstruction only from walk trajectories is surprisingly effective with the proposed data-driven approach. We also note that at this stage there are still obvious ways to drastically improve the performance, first and foremost using a sufficiently large collection of real training data.

% ------------------------------------------------------------------------
% ---------     Repeatability with alternative trajectories      ---------
% ------------------------------------------------------------------------

\subsection{Repeatability with respect to different trajectories}\label{sec:repeatability}

\begin{figure}[t]
  \includegraphics[width=\columnwidth]{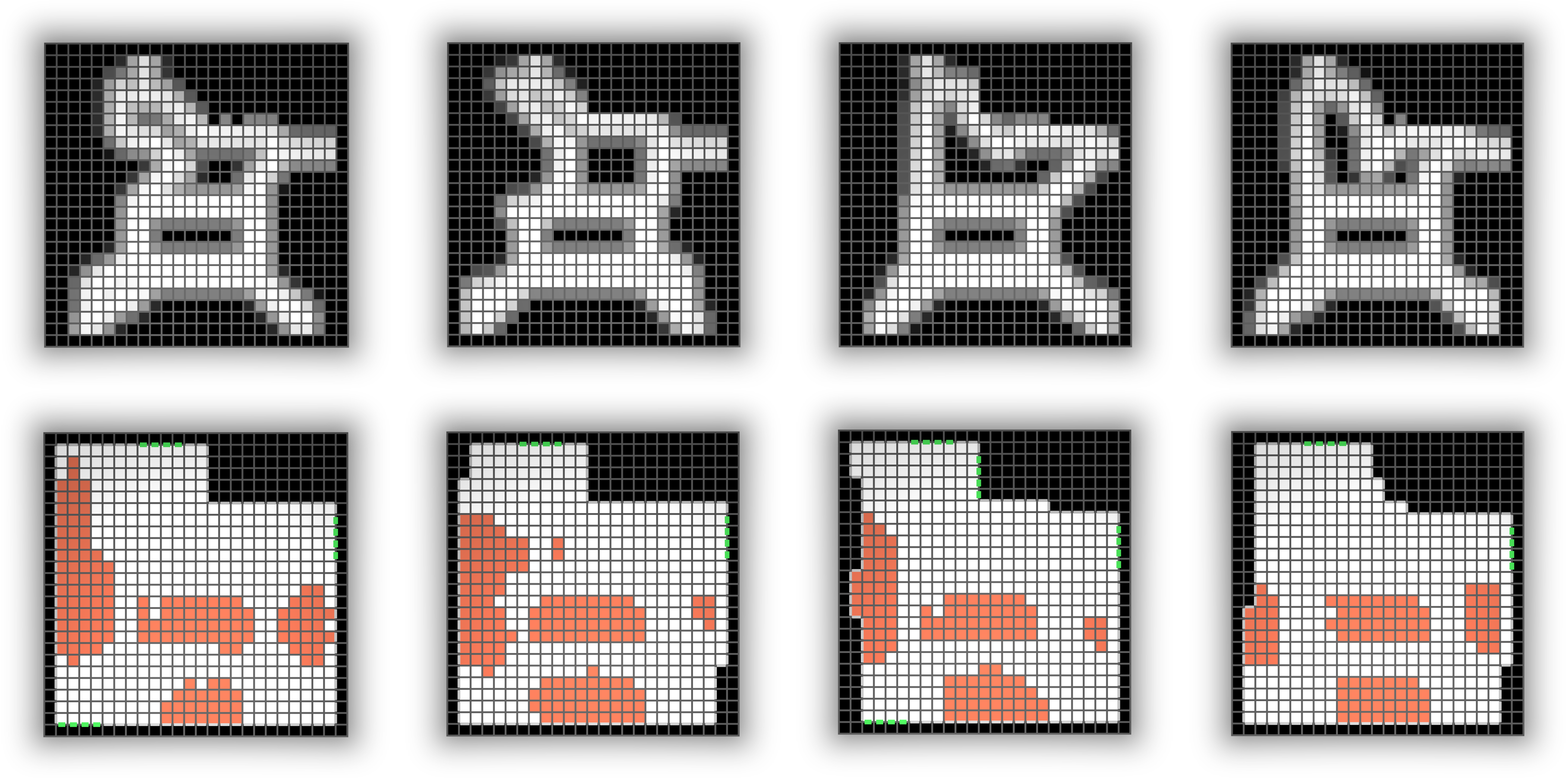}\vspace{-0.5em}
  \caption{Reconstruction repeatability with respect to different trajectories. The same floor plan is reconstructed from 4 different simulated trajectories, leading to fairly consistent results. }\vspace{-2em}
  \label{fig:repeatability}
\end{figure}

We evaluated the consistency of the reconstruction with respect to variations in the input trajectory. To this end, we selected a sample from the Matterport3D test set (model (d) in Figs.~\ref{fig:qualitative_evaluation_floorplans},~\ref{fig:qualitative_evaluation_3D_models}) with several furniture blocks and some degree of irregularity in the footprint, thus being likely to result in non-trivial trajectories. We generated multiple walk trajectories for this sample by running the randomized trajectory simulation with different seeds, and processed the resulting trajectories with our pipeline. 
As shown in Fig.~\ref{fig:repeatability}, the results obtained are generally consistent, with only slight differences in the room footprint and in the shape of the furniture blocks. 

Note that, if several trajectories from different people are available for the same floor plan, the current approach can be adapted to exploit this redundancy, e.g. by converting each trajectory into a separate inverse distance map $\mathcal{I}^{\mathrm{walk}}_\mathcal{C}$ and stacking the resulting maps into a single one, to be used as input to the pipeline.

% ---------------------------------------------------
% ---------       Baseline comparison       ---------
% ---------------------------------------------------

\subsection{Comparison with \emph{pix2pix} baseline}\label{sec:baseline_comparison}

To the best of our knowledge, there exists no other method to generate floor plan models from walk trajectories; hence, a direct comparison is not possible. 
As a baseline, we consider the mapping from a walking trajectory (in inverse distance representation) to a floor plan as a generic image-to-image translation and apply the well-known \emph{pix2pix} method~\cite{Isola:2017:ITC}. In other words, the translation result is a semantic map with labels $\mathbf{out}, \mathbf{in}, \mathbf{furn}, \mathbf{wall}, \mathbf{door}$.
We converted the DFPGs generated from Matterport3D into grayscale images, encoding the five labels with different gray levels; to ensure that the \emph{pix2pix} result only contains valid labels, we quantized the raw outputs to the nearest of those gray levels.
Moreover, to enable the representation of doors in such a way that \emph{pix2pix} can handle them, we enlarged the floor plans to $256\times256$ pixels ($4\times$ the one used for \emph{Walk2map}) and rendered walls and doors with a thickness of $4$ pixels.

\begin{figure}[t]
  \includegraphics[width=\columnwidth]{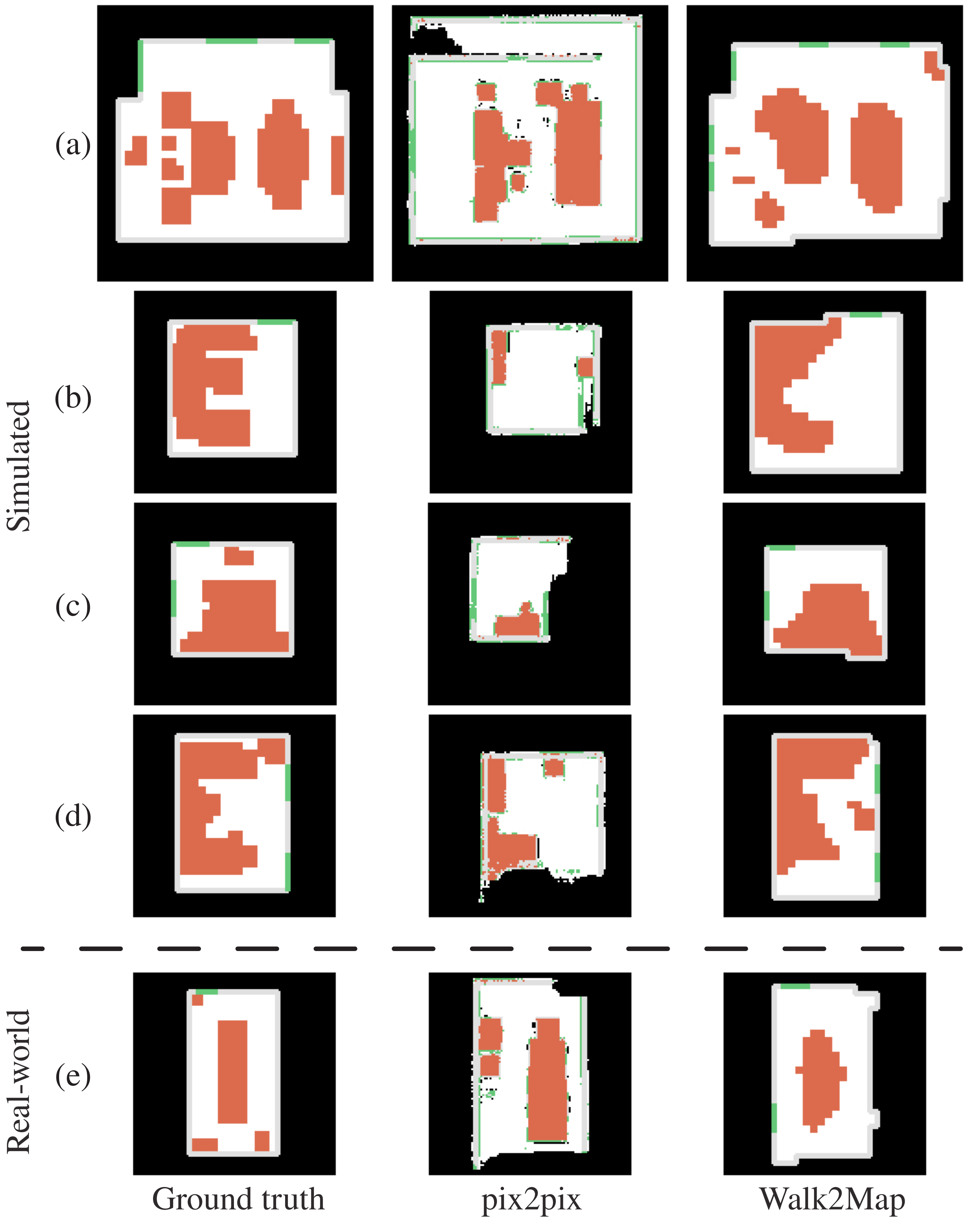}
  \caption{Qualitative examples of the \emph{pix2pix} baseline for Matterport3D. From left to right: ground truth; \emph{pix2pix} prediction; \emph{Walk2map} output. The bottom row refers to a real-world input. (Rendered in the image-based representation of \emph{pix2pix} for visual comparison, with color-coding)}\vspace{-2em}
  \label{fig:baseline_comparison}
\end{figure}

% standard pix2pix model, trained for 200 epochs (default)
We then performed a quantitative evaluation of \emph{pix2pix}, similar to the one described in Secs.~\ref{sec:eval_synthetic} and~\ref{sec:eval_real-world}. The \emph{pix2pix} outputs are very noisy, including long, thin lines of spurious labels (see Fig.~\ref{fig:baseline_comparison}). Therefore, we filter them when extracting doors: we only consider connected regions whose (axis-aligned) bounding boxes have a minimum size of $4\times8$, that is, whose dimensions are at least equal to that of a rasterized door; and whose width is at least equal to $0.5m$ in metric units. As done for our pipeline, we count a predicted element (i.e., an interior/furniture cell or a door) as correctly detected if a corresponding predicted element lies at a distance $\leq 0.25m$; for doors, this distance is defined as the \emph{minimum} distance between the constituting line segments.

As shown in Tab.~\ref{tab:quantitative_results_all} (columns `pix2pix'), this baseline approach performs consistently worse than \emph{Walk2Map} for all floor plan elements, both on simulated and on real-world test data.
Remarkably, despite the domain gap between training and test data, the performance achieved by \emph{Walk2Map} on real-world trajectories is clearly superior to the performance of the baseline trained \emph{and} tested on simulated data.

The limits of the baseline are even more evident in a qualitative visual analysis (Fig.~\ref{fig:baseline_comparison}). The predicted floor plans exhibit some structural resemblance with the ground truth, but individual elements are misplaced and not assembled in a meaningful way. Furniture blocks are generally poorly localized and reconstructed (b, d), doors are often missing (d), and even the basic footprint often deviates significantly from the ground truth (c).
Moreover, the predictions are relatively noisy and not straightforward to clean up, across all labels. 
The most problematic drawback of naive image-to-image translation is its complete ignorance of the inherent structural properties of interiors. For instance, walls do not form closed loops (all examples) and doors appear outside of walls (b).
In contrast, \emph{Walk2map} avoids such problems by breaking up the overall task into simpler sub-tasks, such that basic structural knowledge can be hardwired, leading to floor plans that are structurally sound by construction (e.g., doors can only ever be placed in walls).

% ---------------------------------------------
% ---------       Failure cases       ---------
% ---------------------------------------------

\subsection{Failure cases}\label{sec:failure_cases}

\begin{figure}[t]
  \includegraphics[width=\columnwidth]{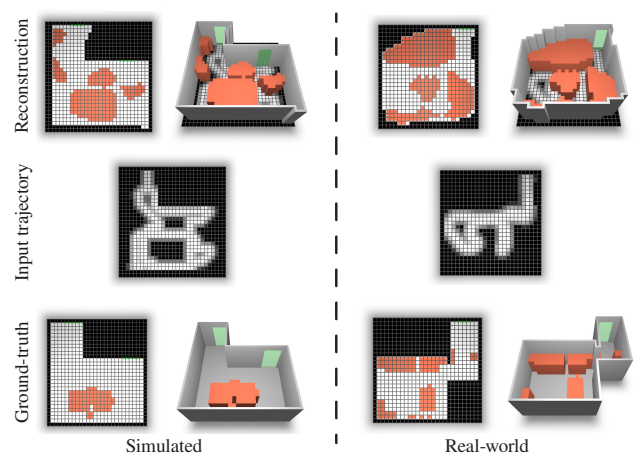}
  \caption{Failure cases. The intrinsic ambiguity of the  walk trajectories can elicit spurious furniture (left), with the hallucinated furniture possibly displacing the walls (right). Such failure cases occur for both simulated (left) and real-world (right) inputs.}\vspace{-2.5em}
  \label{fig:failure_cases}
\end{figure}

Due to our structured approach, we have not observed any \emph{complete} failure in our experiments where the predicted floor plan was inconsistent.
Nonetheless, deriving the locations of walls and furniture only from the sparse evidence for free space that walk trajectories provide inevitably encounters some ambiguity. 
This is the main source of errors in our predicted floor plans, as shown in Fig.~\ref{fig:failure_cases}. The trajectory shown to the left contains several loops, one of which is erroneously interpreted as an indication of furniture; this is also the case for the smaller loop shown in Figs.~\ref{fig:qualitative_evaluation_floorplans_rw}(f) and~\ref{fig:qualitative_evaluation_3D_models_rw}(f), which in fact is not caused by the presence of objects.
In some cases, walls are displaced outwards and the resulting extra space is filled with furniture, as for the real-world trajectory shown in Fig.~\ref{fig:failure_cases}. 

Some furniture configurations are inherently ambiguous and cannot be predicted only from the trajectory. For instance, the large furniture block shown in Figs.~\ref{fig:qualitative_evaluation_floorplans_rw}(f) and~\ref{fig:qualitative_evaluation_3D_models_rw}(f) corresponds to a group of desks arranged in a closed loop; the free space in the middle is not accessible, while obviously there are also furniture blocks without such a hole. Hence, neither the trajectory evidence nor the learned prior can distinguish the two cases.

In fact, there are many more such ambiguities: for instance, it is in our setting a priori impossible to determine the thickness of a wall-to-wall closet. The fact that relatively correct reconstruction are possible at all in most cases confirms the fundamental intuition behind our work: that a trajectory not only proves that the immediate locations it traverses are free space, but carries a lot more information due to contextual cues like proportions, co-occurrences etc.
We believe that, so far, our system has picked up only a fraction of these subtle correlations, and just showing it on a larger and more comprehensive training set will further improve the predicted floor plans quite significantly.
While this extension is conceptually trivial, it is a formidable task to create a large-scale, annotated dataset of real-world floor plans with associated walk trajectories.
Still, there is a limit to what can be derived only from trajectories. Fortunately, it seems entirely possible to augment them with other types of information, including velocities or sparse distances from a portable range sensor.

%-------------------------------------------------------------------------

%-------------------------------------------------------------------------

\section{Conclusions and Future Work}
\label{sec:conclusions_future_work}

We have explored the possibility to automatically derive 2D floor plans only from walk trajectories and have presented \emph{Walk2Map}, the first computational model for this task. The output floor plans include three main types of elements, namely the 2D footprint of the interior space, the location of doors and the furniture layout. These elements are obtained with dedicated neural networks tailored to the individual tasks, which are assembled sequentially into a multistage pipeline. 
Experiments on synthetic and real data confirm that \emph{Walk2Map} can indeed recover valid floor plans that are surprisingly close to the true layout.

Being -- to our knowledge -- the first shot at a seriously ill-posed problem, our current implementation has a number of limitations.
First, due to the limited availability of real-world training data, we had to train entirely on simulated trajectories. 
Despite a successful proof-of-concept evaluation on a small real-world test set, a lot more real data will be needed to employ \emph{Walk2Map} ``in the wild''. 
We note that the lack of data also likely affects the current model's capacity to represent many more intricate patterns in the data, since we had to restrict the complexity of the networks to facilitate training with small datasets.

We see several exciting ways to extend \emph{Walk2Map}. First of all, the input walk trajectories could be augmented with additional information, including height, velocity and timestamps. This could potentially enable the extraction of higher-level semantic information, such as room type, different categories of furniture, the location of windows on wall boundaries, and the segmentation of multi-room environments.
Yet again, these extensions depend on the availability of a large enough training dataset containing both trajectories with all desired observations and associated ground truth floor plans with suitable annotations. Recording such a dataset stands as an important future endeavor. 

Other extensions do not depend on the availability of additional data. First of all, one can obtain more realistic indoor scenes from our predicted floor plans by replacing each connected region labeled as furniture with a 3D model obtained from a furniture database, e.g., by selecting the model whose bounding volume best matches the one of the region.
Moreover, it would be interesting to perform a probabilistic floor plan extraction by exploiting per-cell confidence values in the reconstruction process.

We find it remarkable how much information about the environment is apparently contained in the ego-motion of an agent traversing the scene, without any external sensing or mapping. We believe this opens up an entire new research direction and could play an important role in future systems for systematic, large-scale mapping of indoor spaces.

%-------------------------------------------------------------------------

\begin{spacing}{1.0}
\begin{footnotesize}
  \noindent\textbf{Acknowledgments.} This work was funded by the UZH `Forschungskredit' program (grant no. FK-18-022), by a SNSF Scientific Exchange grant (no. 184583) and by the Hasler Foundation (project no. 20035). We also acknowledge the support of the EU MSCA-ITN project EVOCATION (grant agreement 813170). \vspace{-1em}
\end{footnotesize}
\end{spacing}

%-------------------------------------------------------------------------

% bibtex
\bibliographystyle{eg-alpha-doi}  
\bibliography{bibliography}

% biblatex with biber
%\printbibliography                

%-------------------------------------------------------------------------

\end{document}